\documentclass{article}
\usepackage{arxiv}
\usepackage{color,soul}

\usepackage[utf8]{inputenc} % allow utf-8 input
\usepackage[T1]{fontenc}    % use 8-bit T1 fonts
\usepackage{hyperref}       % hyperlinks
\usepackage{url}            % simple URL typesetting
\usepackage{booktabs}       % professional-quality tables
\usepackage{amsfonts}       % blackboard math symbols
\usepackage{nicefrac}       % compact symbols for 1/2, etc.
\usepackage{microtype}      % microtypography
\usepackage{float}
\usepackage{amsmath,amsfonts,amssymb,graphicx}
\usepackage{natbib}
\usepackage{comment}

\title{Different Spectral Representations in Optimized Artificial Neural Networks and Brains}

\author{%
  Richard C. Gerum\\
  Department of Physics and Astronomy\\
  York University University\\
  Toronto, ON, Canada \\
  \texttt{richard.gerum@protonmail.com} \\
  \And
  Cassidy Pirlot \\
  Departments of Computing Science and Psychology\\
  University of Alberta\\
  Edmonton, AB, Canada \\
  \texttt{pirlot@ualberta.ca} \\
  \AND
  Alona Fyshe \\
  Departments of Computing Science and Psychology\\
  University of Alberta \\
  Edmonton, AB, Canada \\
  \texttt{alona@ualberta.ca} \\
  \And
  Joel Zylberberg\\
  Department of Physics and Astronomy\\
  York University University\\
  Toronto, ON, Canada \\
  \texttt{joelzy@yorku.ca} \\
}

\begin{document}
\maketitle

\begin{abstract}
Recent studies suggest that artificial neural networks (ANNs) that match the spectral properties of the mammalian visual cortex---namely, the $\sim 1/n$ eigenspectrum of the covariance matrix of neural activities---achieve higher object recognition performance and robustness to adversarial attacks than those that do not. To our knowledge, however, no previous work systematically explored how modifying the ANN's spectral properties affects performance. To fill this gap, we performed a systematic search over spectral regularizers, forcing the ANN's eigenspectrum to follow $1/n^\alpha$ power laws with different exponents $\alpha$. We found that larger powers (around 2--3) lead to better validation accuracy and more robustness to adversarial attacks on dense networks. This surprising finding applied to both shallow and deep networks and it overturns the notion that the brain-like spectrum (corresponding to $\alpha \sim 1$) always optimizes ANN performance and/or robustness. For convolutional networks, the best $\alpha$ values depend on the task complexity and evaluation metric: lower $\alpha$ values optimized validation accuracy and robustness to adversarial attack for networks performing a simple object recognition task (categorizing MNIST images of handwritten digits); for a more complex task (categorizing CIFAR-10 natural images), we found that lower $\alpha$ values optimized validation accuracy whereas higher $\alpha$ values optimized adversarial robustness. These results have two main implications. First, they cast doubt on the notion that brain-like spectral properties ($\alpha \sim 1$) \emph{always} optimize ANN performance. Second, they demonstrate the potential for fine-tuned spectral regularizers to optimize a chosen design metric, i.e., accuracy and/or robustness.
\end{abstract}

\section{Introduction}
Artificial neural networks (ANNs) trained to perform object recognition tasks learn representations of input images that bear a surprising resemblance to the ones observed within the mammalian visual cortex \citep{yamins2014performance,yamins2016using,khaligh2014deep}. Excitingly, this resemblance suggests that common principles might underlie both the task-trained ANNs and the mammalian visual system (MVS) \citep{richards2019deep}, and highlights the potential for advances in machine learning and in neuroscience to reinforce each other. Building on these successes, two recent studies \citep{Nassar2020,Kong2022} asked whether the spectral properties of the visual representations in mammalian visual cortex---namely, the $\sim 1/n$ eigenspectrum of the covariance matrix of neural activities \citep{Stringer2019}---might have functional benefit for ANNs. First, \citet{Kong2022} showed that networks which were more robust to adversarial attacks, had eigenspectra closer to those observed in the brain. Second, and more directly, \citet{Nassar2020} trained ANNs with a regularizer that forced them to match the $\sim 1/n$ eigenspectrum seen in neuroscience experiments and found that the resultant ANNs were more robust to adversarial attacks than were ANNs trained without the spectral regularizer. These results suggest that the $\sim 1/n$ eigenspectrum might be among the lessons learned from the MVS that can enhance ANN function\footnote{Another recent study by \citet{Ghosh2022} also concluded that $\sim 1/n$ eigenspectra are associated with improved performance in ANNs. We discuss the relationship between our findings and this previous result in detail in our Discussion section.}, but they also open the question of whether other spectral regularizers (i.e., ones that generate ANNs with eigenspectra following power law $1/n^\alpha$ with power $\alpha$ significantly different than 1) might be even better.

To address that outstanding question, we performed a systematic search over spectral regularizers, characterized by their power law exponents $\alpha$, and found that, for dense networks, larger powers (around 2--3) lead to better validation accuracy and to better robustness to adversarial attacks. This surprising finding persisted for both shallow and deep dense networks. For convolutional networks, the best $\alpha$ values depended on the evaluation metric and task. For a simpler object recognition task (categorizing MNIST images of handwritten digits), validation accuracy and adversarial robustness were best for smaller values of $\alpha$ (around 1). For a more complex task (categorizing natural images from the CIFAR 10 dataset), validation accuracy was best for smaller $\alpha$ values but robustness to adversarial attacks was best for larger values of $\alpha$ (around 2--3).

These results have two main implications. First, they suggest that low dimensional image representations\footnote{higher exponents $\alpha$ imply lower dimensional representations} might be a guiding principle for developing robust ANNs for complex visual tasks, and highlight the potential for spectral regularizers with appropriately-chosen $\alpha$ values to optimize ANNs for different task types. Second, the deviation between the optimal spectral properties of the object-recognition ANNs and that which is observed in the MVS suggests that there could be important deviations between the principles underlying object-recognition ANNs and those that underlie the MVS. For example, perhaps ANNs optimized for tasks other than complex object recognition and/or robustness might be better matches to the MVS \citep{zhuang2021unsupervised,bakhtiari2021functional,christensen2020models}, or perhaps there are fundamental differences between ANNs and the MVS that extend beyond the choice of objective function. We leave the task of speculating about the answers to these interesting questions for future work.

\begin{figure}
    \centering
    \includegraphics{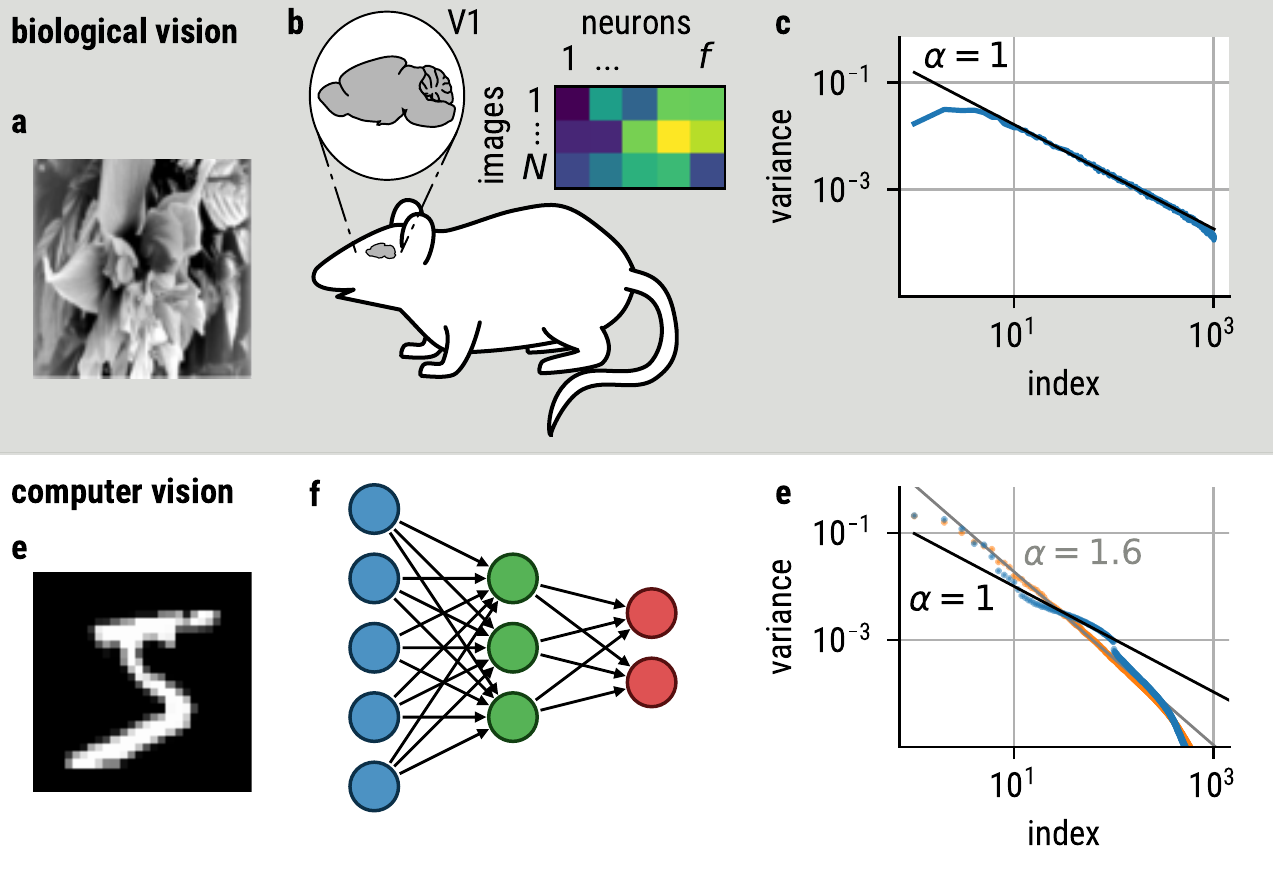}
    \caption{\textbf{a}, Image presented to a mouse. \textbf{b}, From the visual cortex of the mouse the neural representations of $f$ neurons are recorded in response to $N$ images. \textbf{c}, The values of rank-ordered eigenvalues along with a power-law fit (data from \cite{Stringer2019}). \textbf{e}, MNIST digit, presented to a dense artificial neural network \textbf{f}. The activation of the hidden layer to a batch of images can be analysed analogously to the neural recordings to obtain the eigenvalue spectrum \textbf{e}. \textbf{e}, Eigenvalue spectrum for an unregularized network (orange, gray fit) and for a regularized network (blue, black fit).}
    \label{fig:setup}
\end{figure}

\section{Method}
\subsection{Eigenspectrum}
Our study is motivated by observations of the spectral properties of visual cortex. To measure those spectral properties, \cite{Stringer2019} presented images to mice and recorded the neural activity in the primary visual cortex (V1). This yields a partial snapshot of the neural representation of the image (Fig.\ \ref{fig:setup}a,b). After obtaining the activity from $f$ neurons in response to $N$ images, the data can be represented in a matrix $A$.

For this matrix $A$, the matrix $\Sigma$, which is proportional to the data covariance matrix, is calculated from the mean corrected data $\hat{A}$.
\begin{align}
    \hat{A_i} &= A_i - \bar{A} \\
    \Sigma &= \hat{A}^\mathrm{T} \hat{A} 
\end{align}
where $\bar{\cdot}$ denotes the average over $i$.

This matrix $\Sigma$ can be decomposed into its eigenvalues and vectors.
\begin{align}
    \Sigma &= V \Lambda V^\mathrm{T}
\end{align}
where $V$ is an orthonormal matrix of eigenvectors and $\Lambda$ is a diagonal matrix with the (sorted) eigenvalues $\lambda_i$ of $\Sigma$ on the diagonal.

Plotting the these eigenvalues, sorted descending, over their order indices $I_i$, ranging from 1 to the number of eigenvalues $n$, \citet{Stringer2019} saw a decaying power-law with a slope of -1 (Fig.\ \ref{fig:setup}c).

\subsection{Constructing the Spectral regularizer}
The same principle can be leveraged to get the spectral representation of an artificial neural network. When presenting an image (Fig.\ \ref{fig:setup}e) to a network (Fig.\ \ref{fig:setup}f), the activations of units in a hidden layer can be represented by a matrix $A$ for which the eigenvalues $\lambda$ and their rank-ordered indices $I_i$ can be obtained. To estimate the decay of the eigenvalue spectrum, we fit a power-law to the eigenvectors $\lambda_i$ over the indices of sorting $I_i$.

Therefore, we use linear least squares fit to the logarithms of these which has a closed form
\begin{align}
    \hat{\alpha} = \frac{\sum_{i=i_\mathrm{min}}^{i_\mathrm{max}} (\log{I_i} - \left<\log{ I}\right>_i)(\log{\lambda_i} - \left<\log{\lambda}\right>_i)}{
    \sum_{i=i_\mathrm{min}}^{i_\mathrm{max}} (\log{I_i} - \left<\log{I}\right>_i)^2
    }
\end{align}
Where $\left<\cdot\right>_i$ denotes the average in the range $i_\mathrm{min}\leq i \leq i_\mathrm{max}$. The range $[i_\mathrm{min}, i_\mathrm{max}]$ is chosen to capture the linear range of the spectrum.

We now define the value $\alpha$ as the negative of this slope:
\begin{align}
    \alpha = -\hat{\alpha}
\end{align}

We use the absolute deviation from the chosen target alpha value, $\alpha_\mathrm{target}$, as a regularizing loss term with strength $\beta$:

\begin{align}
\label{eq:regularizer}
    \operatorname{loss} = \beta \left|\alpha - \alpha_\mathrm{target}\right|
\end{align}

This spectral loss is differentiable and hence it allows for the gradient of this loss to be used during training.

\subsection{Assessing Adversarial Robustness}
To determine whether this regularization method leads to more robust networks, we train networks with different regularization strengths, $\beta$, and target values, $\alpha_\mathrm{target}$, and then perform two different kinds of common attacks on the trained networks.

The fast gradient sign method (FGSM, \citet{goodfellow2014explaining}) perturbs an input image with pixel values $x^j_i$ and label $y^j$, to one with pixel values $\hat{x}^j_i$:
\begin{align}
    \hat{x}^j_i = x^j_i + \epsilon\operatorname{sgn}\left(\frac{\partial}{\partial x_i} L\left(f(x^j, \Theta), y^j\right)\right)
\end{align}
The perturbation depends on the sign of the derivative of the classification loss function $L(\cdot)$ (in our case the categorical cross entropy loss) with respect to the $i$th input pixel, multiplied with an attack strength $\epsilon$. This perturbation is added to the initial input image $x^j$ to produce the attack image $\hat{x}^j$. To ensure that the attack image $\hat{x}^j$ is still a valid image, values are clipped to the range of $[0,1]$. This perturbation has the effect of modifying the pixel values so as to increase the loss, and hence decrease the likelihood that the network can correctly categorize the image.

The projected gradient descent (PGD, \citet{mkadry2017towards}), takes an iterative approach using gradient ascent to find changes to the input image that maximize the loss function.
\begin{align}
    \hat{x}^{j, n+1}_i = \operatorname*{proj}_{\|x-\hat{x}\|_{\infty}\leq \epsilon}\hat{x}^{j, n}_i + \mu\left(\frac{\partial}{\partial x_i} L\left(f(x^j, \Theta), y^j\right)\right)
\end{align}
with $\hat{x}^{j, n}$, referring to the attack image at the $n$th iteration, $\mu$ the learning rate for the gradient ascent, $\operatorname*{proj}(\cdot)$ to projecting the change to a maximum change of $\epsilon$, the attack strength, per pixel ($\|\cdot\|_\infty$ refers to the infinity norm). We use a learning rate $\mu$ of 0.01 and 40 iterations.

\section{Results}
\label{sec:results}
\begin{figure}[htbp]
    \centering
    \includegraphics{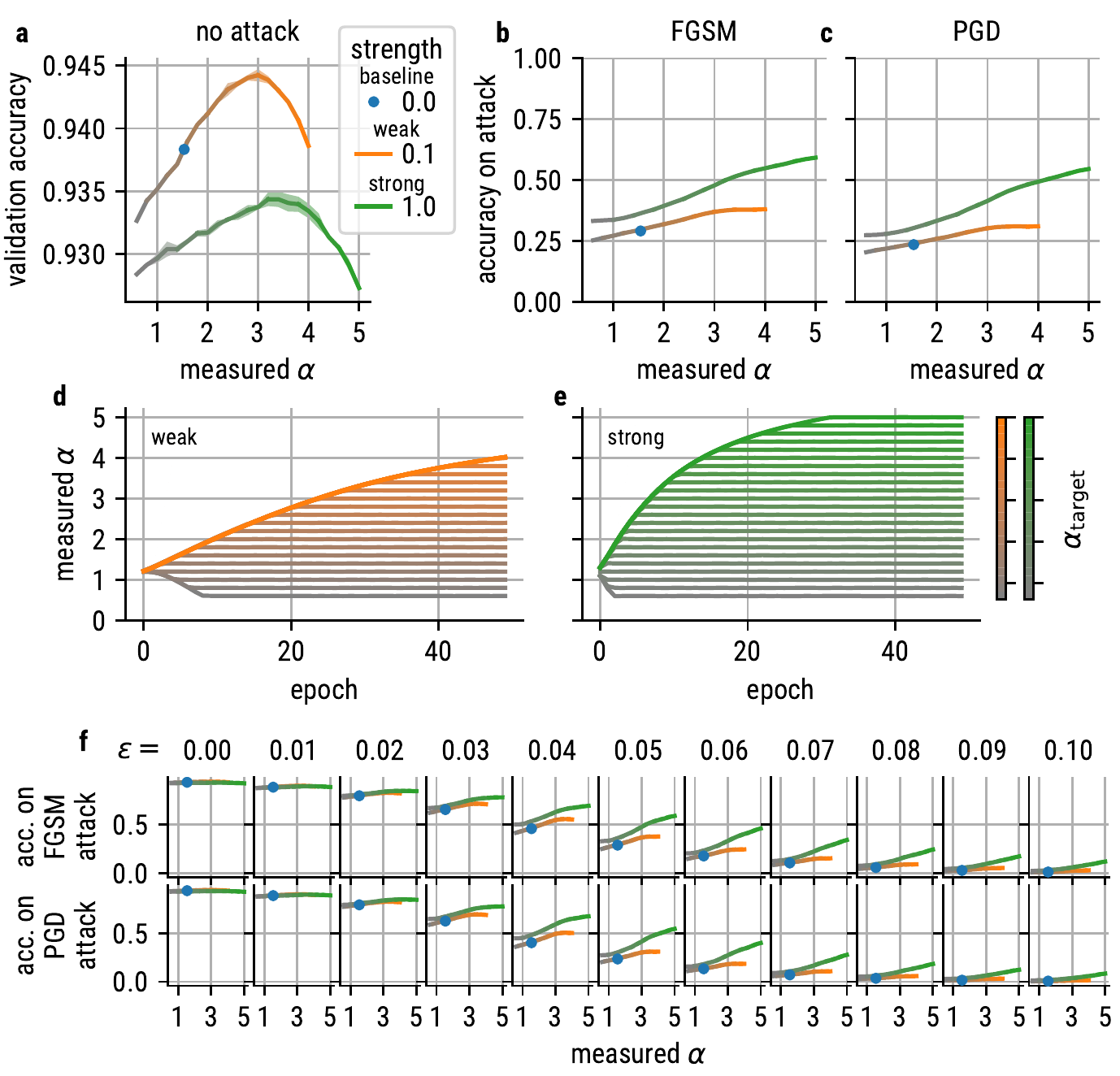}
    \caption{\textbf{a}, Validation accuracy over spectral slope, $\alpha$, after training a shallow perceptron with one 2000 unit hidden layer on MNIST for 50 epochs with spectral regularization for different regularization strengths $\beta$ ("baseline": no regularization, blue; "weak": $\beta=0.1$, orange; "strong": $\beta=1.0$, green; color saturation shows target $\alpha$, see color bars). Adversarial attacks with \textbf{b} FGSM and \textbf{c} PGD with a strength $\epsilon=0.05$.
    \textbf{d},\textbf{e}, Measured $\alpha$ value during training for different target $\alpha$ values (bright orange/green $\alpha_\mathrm{target}=5$, gray $\alpha_\mathrm{target}=0.6$) for weak regularization strength (orange, $\beta=0.1$) and strong regularization (green, $\beta=1.0$). \textbf{f}, Robustness against attacks for different attack strengths $\epsilon$ show comparable results to \textbf{a}--\textbf{c}.}
    \label{fig:convergence_alpha}
\end{figure}

\begin{figure}[htbp]
    \centering
    \includegraphics{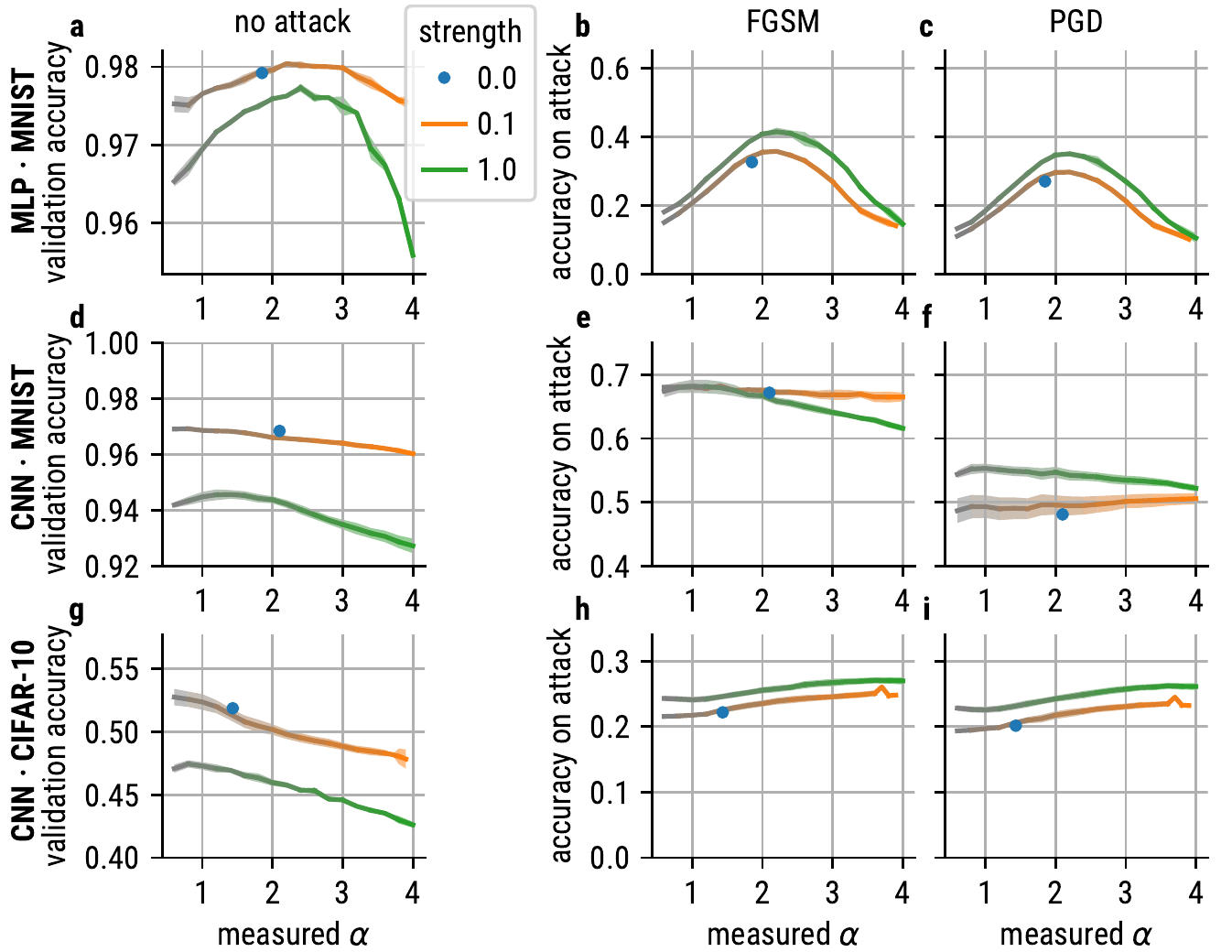}
    \caption{Validation accuracy and adversarial robustness for different architectures. \textbf{a},\textbf{b},\textbf{c}, Multilayer perceptron with 3 hidden layers trained on MNIST. \textbf{a}, Validation accuracy is higher than baseline (blue) for alpha of 2.4 and weak regularization strength (orange) but lower for stronger regularization (green). \textbf{b}, FGSM and \textbf{c} PGD with a strength $\epsilon=0.05$ show similar robustnes for $\alpha$ values of 2.1. \textbf{d},\textbf{e},\textbf{f}, Convolutional neural network trained on MNIST. \textbf{d}, Validation accuracy cannot be improved against baseline and only shows a clear maximum for strong regularization. \textbf{e}, FGSM and PGD \textbf{f} attacks with a strength $\epsilon=0.1$ show only strong differences for regularization strength for PGD and do not have a pronounced maximum. \textbf{g},\textbf{h},\textbf{i}, Convolutional neural network trained on CIFAR-10. \textbf{g}, Validation accuracy is strongest for weak regularization and $\alpha$ values smaller than one. \textbf{h}, FGSM and PGD \textbf{i} attacks with a strength $\epsilon=0.01$ show that strong regularization and $\alpha$ values around 3 improve robustness.}
    \label{fig:architectures}
\end{figure}

\subsection[{Optimizing alpha to Improve Accuracy and Robustness}]{Optimizing $\alpha$ to Improve Accuracy and Robustness}

We first trained a shallow densely connected network to categorize MNIST images of handwritten digits: our later experiments explore deeper networks, convolutional architectures, and more complex tasks. This first network had one hidden layer with 2000 units with tanh activations followed by a softmax output layer. We train with a batch size of 2500 for 50 epochs using the Adam optimizer with a low learning rate of $10^{-4}$. The network architectures and hyperparameters were chosen to be comparable to the experiments that \citet{Nassar2020} performed.

During training, the representation in the hidden layer was regularized with the spectral regularizer (Eq.\ \ref{eq:regularizer}), which encourages the eigenspectrum of the covariance matrix to follow a power-law with the given slope $\alpha_\mathrm{target}$.

Validation accuracy (without any attack) depended only slightly on the regularization strength and $\alpha_\mathrm{target}$ (between 0.93 and 0.95, Fig.\ \ref{fig:convergence_alpha}a), and regularization was only beneficial compared to the unregularized network for small regularization strengths ($\beta=0.1$) and alpha values higher than 1.5. The optimal $\alpha$ was around 3. For stronger regularized networks ($\beta = 1.0$), no $\alpha_\mathrm{target}$ value yielded a higher accuracy than the unregularized network.

However, testing the network for robustness against adversarial attacks, led to quite a different picture. When $\alpha_\mathrm{target}$ was such that the weakly regularized network had the same $\alpha$ value as the baseline network (with no regularization), both the weakly regularized and the unregularized baseline network had the same robustness to adversarial attacks (Fig.\ \ref{fig:convergence_alpha}b,c). At the same time, higher $\alpha_\mathrm{target}$ values, leading to higher $\alpha$ values in the trained networks, led to greater robustness.

For stronger regularization (green curve in Fig.\ \ref{fig:convergence_alpha}a--c), we generally observed greater robustness to adversarial attacks, and once again found that higher $\alpha$ values were associated with greater robustness. Stronger regularisation differed from the weaker regularisation in reaching the target alpha value earlier in training (Fig.\ \ref{fig:convergence_alpha}d,e).

These findings were qualitatively unchanged when using a PGD attack instead of an FGSM attack (Fig.\ \ref{fig:convergence_alpha}c). They also do not depend on the exact strength of the attack (Fig.\ \ref{fig:convergence_alpha}f), as qualitatively similar results were observed for different attack strengths $\epsilon$.

Overall, these results highlight that spectral regularization can improve both validation accuracy and adversarial robustness, and that the optimal $\alpha$ values can be much larger than those that are observed in the mammalian visual system.

\subsection{Deeper and Convolutional Architectures}
To investigate how much our results depend on the network architecture and/or task complexity, we trained deeper dense networks and convolutional networks to categorize MNIST images of handwritten digits. We separately trained the convolutional networks to categorize CIFAR-10 natural images (We do not include results of dense networks on the CIFAR-10 dataset because performance was too low.) These network architectures were chosen to match the ones used in a recent study of spectral regularizers \citep{Nassar2020}.

The specific network architectures were as follows. Our deeper dense network had three hidden layers each with 1000 neurons (tanh activation) followed by a 10 unit softmax output layer (Fig.\ \ref{fig:architectures}a,b,c). This network was trained on MNIST. Our convolutional network had two convolutional layers (16 filters, 3x3 kernel, and 32 units 3x3 kernel, tanh activation) followed by a dense softmax output layer, which we trained on either MNIST (Fig.\ \ref{fig:architectures}e,f,g) or CIFAR10 (Fig.\ \ref{fig:architectures}g,h,i) for 200 epochs. 

For each trained network, we report validation accuracy, and we report robustness against attacks of a strength $\epsilon$. This attack strength was chosen so as to lower the accuracy to about 50\% of the validation accuracy. The required attack strength was different for the different networks (see figure caption). We include the results for the whole range of attack strengths we explored in the Supplementary Material (Fig.\ S1).

Training the deeper dense network on the MNIST dataset showed qualitatively the same results as we found with the shallower dense network (Fig.\ \ref{fig:convergence_alpha}), with an optimal $\alpha$ of $\sim 3$. Once again, weak regularization showed the best validation accuracy (Fig.\ \ref{fig:architectures}a). Stronger regularization led to more robustness against adversarial attacks, and we observe that $\alpha \sim 2$ provides the best protection against attacks (Fig.\ \ref{fig:architectures}b,c).

When training the convolutional network on the MNIST dataset, there was no clear peak defining the optimal $\alpha$ value, and strong regularization only provided an improvement on PDG attacks. Nevertheless, lower $\alpha$ values were generally associated with better validation accuracy in this case.

Finally, training the CNN on the MNIST dataset showed a surprising plateau for values of $\alpha < 1$ for validation accuracy. For robustness against attacks, we observe a broad peak around $\alpha \sim 3$.

These results reinforce our conclusions obtained with the shallow dense network, namely that spectral regularization can improve both validation accuracy and adversarial robustness, and that the optimal $\alpha$ values can be quite different from those that are observed in the mammalian visual system (which are $\sim 1$).

\section{Limitations}
\label{sec:limitations}
While we have demonstrated our main conclusions for a variety of network architectures and for tasks of two different levels of complexity, our study nevertheless has several limitations that must be acknowledged.

First, we have only investigated network architectures of relatively small size. State of the art networks often use far larger architectures which might benefit from different $\alpha$ values than the ones observed here. Also while the architectures we studied here were limited to dense and convolutional networks, there are other relevant architectures that were not included. For example, transformer networks might have an entirely different optimal $\alpha$ value, and the results of \citet{Ghosh2022} suggest that very low $\alpha$ values (much less than 1) might be optimal for these architectures. We leave the test of this hypothesis for future work.

Second, we limited our investigation to image classification tasks, in part to stay close to the work of \citet{Stringer2019} that reported findings on the mammalian visual system. But even for image based tasks, classification is not the only relevant task: image transformations or video prediction tasks \citep{bakhtiari2021functional} could lead to different preferred $\alpha$ values.

Finally, while our work is inspired by the properties of the mammalian visual system, we have investigated only feedforward neural networks here. While this choice is consistent with much previous work in this area 
\citep{Safarani202106a9d51e,Nassar2020,Federer2020,christensen2020models,khaligh2014deep,yamins2014performance}, 
it nevertheless has importance deviation from the recurrent structure of the brain's neural circuits. This discrepancy might be important to consider when observing that ANNs show different optimal representations compared to those found in the brain.

\section{Discussion}
We investigated how the dimensionality of intermediate representation affects the performance and adversarial robustness of dense and convolutional neural networks trained on MNIST and CIFAR-10 object recognition tasks. We performed this investigation by adjusting the power law exponent of of the eigenspectrum of the covariance matrix of unit activations with a regularizer: higher power law exponents correspond to lower dimensionality and vice versa. Previous results and theoretical derivations from \citet{Stringer2019}, \citet{Nassar2020}, or \citet{Ghosh2022} suggested that a power-law exponent of $\alpha \sim 1$ would be the optimal representation because that corresponds to the highest-dimensional representation that is still smooth.

\citet{Ghosh2022} studied networks trained without spectral regularization and found that the best-performing networks (in terms of categorization performance) had $\alpha \sim 1$. The difference between these findings and ours could be attributed to the difference between letting the networks train and the seeing what $\alpha$ value is associated with good performance (the approach of \citet{Ghosh2022}) vs using regularizing to change the $\alpha$ value and then asking which target $\alpha$ is best (our approach). In other words, networks that learn to do the task well might converge to $\alpha \sim 1$ even if forcing $\alpha \to 1$ does not guarantee better performance. In that sense, $\alpha \sim 1$ may not be \emph{causally} linked to improved performance in neural networks. 

Given that \citet{Stringer2019} presented insightful theoretical arguments for why optimal representations should have $\alpha \sim 1$, it is natural to ask why our findings deviate so strongly from that value. Recall that \citet{Stringer2019} argue that an $\alpha$ value of $1+2/d$ would lead to a representation with maximal dimensionality while still keeping the representation smooth. Therefore, $\alpha$ values above one (for large input data dimensionality $d$) would be lower dimensional than is optimal. Consistent with notion, \citet{Stringer2019} find that neural data from visual cortex is consistent with $\alpha \sim 1$. For contrast, we find that higher values of $\alpha$ typically improve performance in our ANNs. 

One speculative explanation for this discrepancy is that, while the brain is optimized for a wide variety of stimuli and tasks, and hence benefits from higher-dimensional representations \citet{rigotti2013importance}, our ANNs were trained on a single image type and task. For these very restricted tasks, it may be more beneficial (in terms of generalization error and adversarial robustness) for the network to extract a small number of task-relevant relevant image features rather than a higher-dimensional representation. This lower dimensionality corresponds to the higher $\alpha$ values we observed. 

The spectral regularization approach to comparing ANN representations to brain representations has the advantage of simplicity and it can be applied in principle to any task, including those for which no brain data is available. At the same time, other, more sophisticated approaches have also been pursued that tailor object recognition neural networks to more directly mimic the brain's visual representations. For example, \citet{Federer2020} regularized convolutional networks to representation similarity matrices that are close to those obtained in the visual cortex when an animal views images, and \citet{Safarani202106a9d51e} which directly trained their network to predict neural responses. This latter approach forces the network to extract sufficient stimulus information to predict the neural responses, and thus encourages a more brain-like representation. Compared to our spectral regularization approach, these other methods require training data consisting of neural activity data and the corresponding images. This limits the application of these methods to cases were such data are available.

\section{Potential negative societal impacts}
\label{sec:impact}
Our work currently only addresses classification of MNIST and CIFAR-10 datasets which are itself quite limited for applications with wide societal impact. Of course this work of using an optimal spectral representation can lead to performance gains in larger and more directly relevant classification tasks and thus also potentially impact society. 
In general, transitioning to networks more robust against adversarial attacks should provide more safety for systems that critically rely on machine learning, e.g.\ autonomous vehicles. At the same time, it might also lead to less defense against possibility malevolent uses of machine learning systems, e.g.\ autocratic surveillance systems.

\section{Data Availability Statement}
\label{sec:data}
There are no primary data in the paper, all code is available at \url{https://github.com/rgerum/spectral-representations} under MIT licence.

\section{Software and Datasets}
\label{sec:software}
Experiments were run with Python \citep{RossumPython} using Keras \citep{chollet2015keras} and the Tensorflow backend \citep{tensorflow2015-whitepaper}. Evaluations used Numpy, Pandas. Results were visualised using Matplotlib \citep{Hunter:2007} and Pylustrator \citep{Gerum2019e}.

We used the MNIST \citep{MNIST} (released under the "Creative Commons Attribution-Share Alike 3.0 license") and the CIFAR-10 \citep{cifar10} datasets (no license specified). Datasets are loaded using the respective Keras functions which include a split into training and validation data.

Experiments were run on a personal computer equipped with Intel® Core™ i9-10980XE CPU @ 3.00GHz × 36 and a NVIDIA GeForce RTX 3080 and on a Shared Cluster with NVIDIA P100 and V100 GPUs.

\bibliographystyle{apalike}
\bibliography{references}

\appendix

\newpage
\renewcommand\thefigure{S\arabic{figure}}
\setcounter{figure}{0} 
\section*{Supplementary Material}

\begin{figure}[hbtp]
    \centering
    \makebox[\textwidth][c]{
    \includegraphics{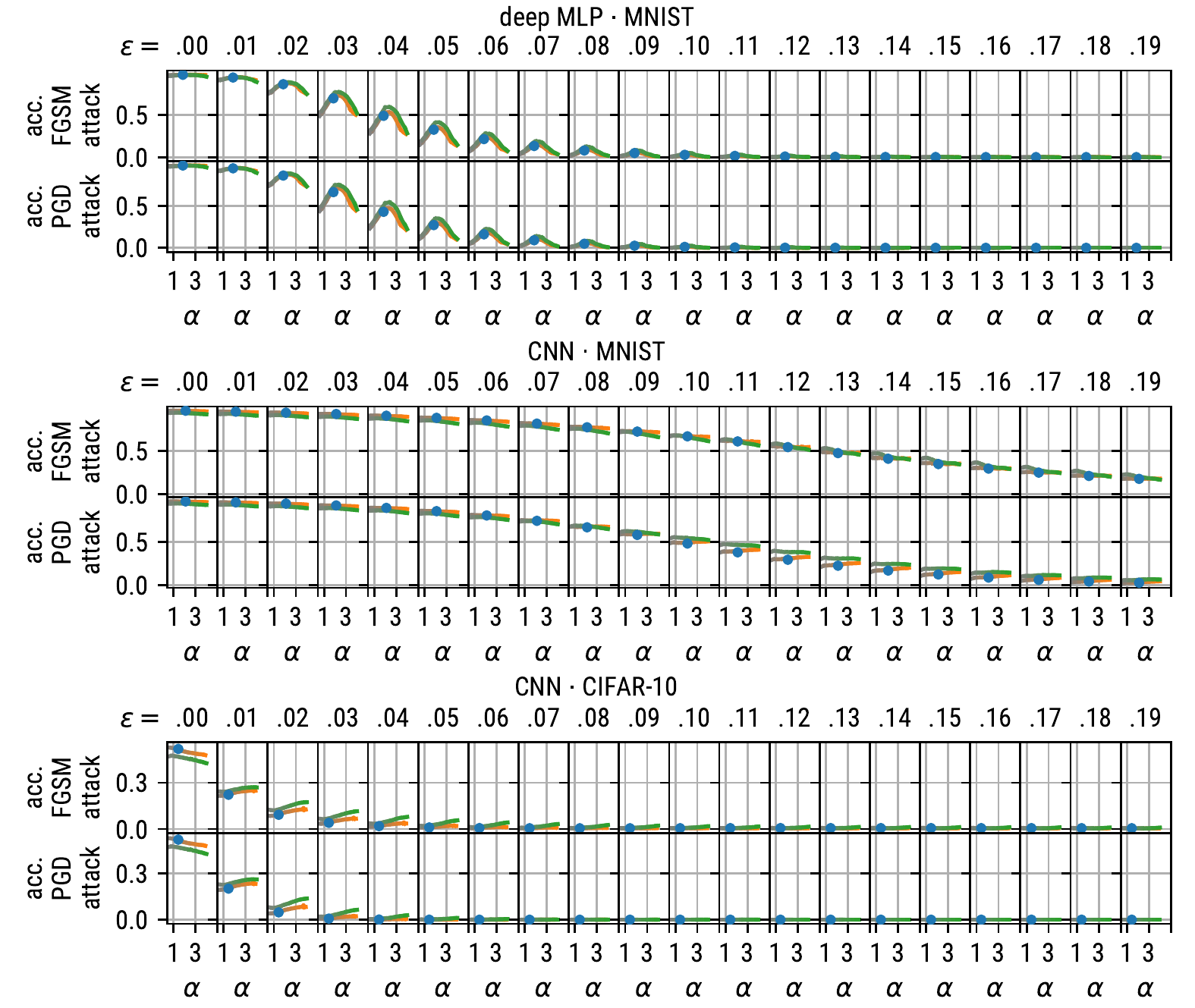}}
    \caption{Robustness against adversarial attacks for the different networks used in the main text: deep MLP, CNN MNIST and CNN CIFAR-10. Upper row: FGSM attack, lower row: PGD attack, columns correspond to different attack strengths. Results for baseline (blue) weak (orange) and strong (green) regularisation.}
    \label{fig:all_attacks}
\end{figure}

\end{document}